%% file: 00_MAIN.tex
\documentclass[letterpaper, 10 pt, conference]{ieeeconf}  
\IEEEoverridecommandlockouts                             
\usepackage{graphics} 
\usepackage{epsfig} 
\usepackage{mathptmx} 
\usepackage{times} 
\usepackage{amsmath} 
\usepackage{amssymb}  
\usepackage{xcolor}
\usepackage{multirow}
\usepackage{algorithm}
\usepackage{algpseudocode}
\usepackage{fancyhdr}
\usepackage{hyperref}
\hypersetup{
    colorlinks=true,
    linkcolor=magenta,
    urlcolor=blue,
    citecolor=red 
}

\title{\LARGE \bf
Integrated Graph Search and Model Predictive Control for Smooth and Efficient Path Planning in Autonomous Vehicles}

\author{Duc-Tien Bui$^{1}$, Ngoc Thinh Nguyen$^{2}$, Hung Duy Nguyen$^{3}$, Dong Bi$^{1}$, Tomislav Mihalj$^{1}$ and Arno Eichberger$^{1}$
\thanks{*This work was supported by ASEAN-UNINET Project}
\thanks{$^{1}$Duc-Tien Bui is with the Institute of Automotive Engineering, Graz University of Technology, 8010 Graz, Austria
        {\tt\small d.t.bui@tugraz.at}}
\thanks{$^{2}$Ngoc Thinh Nguyen is with Drone Engineering, University of Applied Science Kufstein Tirol, Austria
        {\tt\small ngoc-thinh.nguyen@fh-kufstein.ac.at}}       
\thanks{$^{3}$Hung Duy Nguyen is with Automation and Control Institute (ACIN), Vienna University of Technology, Austria
        {\tt\small nguyen@acin.tuwien.ac.at}}
\thanks{$^{1}$Dong Bi is with the Institute of Automotive Engineering, Graz University of Technology, 8010 Graz, Austria 
        {\tt\small dong.bi@tugraz.at}}%
\thanks{$^{1}$Tomislav Mihalj is with the Institute of Automotive Engineering, Graz University of Technology, 8010 Graz, Austria 
        {\tt\small tomislav.mihalj@tugraz.at}}%
\thanks{$^{1}$Arno Eichberger is with the Institute of Automotive Engineering, Graz University of Technology, 8010 Graz, Austria
        {\tt\small arno.eichberger@tugraz.at}}%
}

\begin{document}

\maketitle
\thispagestyle{empty}
\pagestyle{empty}

\begin{abstract}

Path planning is a fundamental component of autonomous vehicles, where achieving safe, comfortable, and dynamically feasible paths while ensuring computational efficiency remains a significant challenge.
This paper presents a sequential path planning framework in which a rough path obtained from graph search is explicitly exploited to guide a Model Predictive Control (MPC)-based path refinement.
A rough path is first obtained via Dijkstra search on a discretized grid and is then used to construct a spatially varying convex lateral safety corridor that explicitly captures obstacle avoidance constraints, transforming discrete obstacle avoidance decisions into continuous feasibility constraints for optimization.
Within this corridor, an MPC problem is formulated to refine the path, enabling efficient optimization while maintaining path smoothness by penalizing the third-order spatial derivative of the lateral offset over a prediction horizon.
The proposed algorithm is evaluated in multiple overtaking scenarios on both straight and curved roads, including cases with single and multiple target vehicles, using high-fidelity environment simulations (i.e., CarMaker).
Compared with the previous study, which used polynomial fitting and a quadratic programming method, the proposed approach consistently achieves lower lateral acceleration, curvature, and jerk while reducing computational cost by 28.08\% on straight roads and 29.52\% on curved roads.
These results demonstrate that exploiting graph-search structure within an MPC formulation provides an effective balance between path smoothness and computational efficiency for autonomous vehicles in structured driving environments.

\end{abstract}
\input{01_INTRO}

\input{02_New_Method}

\input{03_Simulation_results}
\input{04_00.Discussion}
\input{05_CONCLUSION}
{\color{blue}
\addtolength{\textheight}{-12cm}  
}

\bibliographystyle{ieeetr}

\bibliography{Refers}

\end{document}

%% file: 01_INTRO.tex
\section{ INTRODUCTION}
The development of automated vehicles (AVs) has witnessed significant progress in recent years. 
As a core component of AVs, path planning is responsible for generating safe and dynamically feasible paths that guide the vehicle through complex environments while meeting requirements for passenger comfort and computational efficiency 
\cite{abdallaoui2022thorough, reda2024path}.

Path planning can be divided into global and local levels \cite{reda2024path}, \cite{meng2019decoupled},  where global path planning creates a route from the starting point to the ending point and local path planning generates a concrete segment satisfying constraints, adapting to the changing dynamic environment \cite{liu2023path}, \cite{zhang2024intelligent}.
Both layers work cooperatively: the global planner provides macroscopic guidance, whereas the local planner ensures safety, feasibility, and smoothness in execution. 
Despite extensive research, key challenges in path planning remain in achieving smooth path profiles, computation cost and maintaining robustness across diverse driving scenarios \cite{meng2019decoupled}.

To address these challenges, traditional algorithms such as Dijkstra’s, A*, Artificial Potential Field (APF) and their variants have laid the theoretical foundation for early path planning \cite{hongyu2018improved, hart1968formal, yijing2018local, rachmawati2020analysis, nguyen2023risk}. Dijkstra’s algorithm or A* \cite{rachmawati2020analysis} are employed to find the shortest path and the APF method \cite{nguyen2023risk} simulates attractive and repulsive forces to guide a car toward its goal while avoiding obstacles \cite{gonzalez2015review}. Although these methods are computationally efficient and easy to implement, they often suffer from local minima, limited adaptability, and difficulty handling dynamic obstacles \cite{abdallaoui2022thorough}, \cite{reda2024path}, \cite{gonzalez2015review}.

To overcome the limitations of those algorithms, optimization-based methods are developed. These methods formulate the path or trajectory as a constrained linear programming problem, directly minimizing cost functions related to smoothness, safety, and comfort. Representative techniques include Sequential Quadratic Programming \cite{lim2018hierarchical}, Convex Feasible Set \cite{liu2017convex}, Constrained Iterative LQR \cite{chen2017constrained} and MPC \cite{11404227, nezami2024safe}, which iteratively solve convex subproblems for faster convergence. While capable of producing smooth, feasible trajectories, these methods are sensitive to initial conditions and may converge to local optima, particularly in highly dynamic environments \cite{liu2023path}, \cite{katrakazas2015real}.

Additionally, Sampling-Based Planning (SBP) is represented by Rapidly-exploring Random Tree (RRT) and its variants find the path between the start and endpoints with random tree-like branches \cite{gonzalez2015review}, \cite{zhang2023bi}. The paths are guaranteed to be found if given enough run-time. However, SBP approaches have several limitations: produce jerky paths, lack explicit geometric optimization, and require post-processing to ensure smoothness and dynamic feasibility \cite{reda2024path}, \cite{gonzalez2015review}, \cite{zhou2022review}.

In addition, Dynamic programming (DP) provides an efficient approach to addressing the shortest path planning problem. By breaking down the overall task into smaller, more manageable subproblems, it constructs a globally optimal path through the combination of their optimal solutions \cite{9211493}. In real-world implementations, Baidu Apollo employs DP within the EM and lattice planning frameworks, both of which are well-established in prior research \cite{fan2018baidu}. 
The authors in \cite{zhang2024intelligent}, \cite{meng2019decoupled}, \cite{9590740}  proposed a DP-based technique that discretizes the continuous state space combine with some constraints to achieve globally optimal results and improve path smoothness. 
Li et al. \cite{li2023quantitative} employed DP in combination with QP for path planning and analyzed the impact of key parameters on trajectory planning in critical scenarios. 
However, when the resolution increases, computational costs of DP grow exponentially, making execution time challenging. 

These above articles indicate that the development of a path planning capable of the optimality, real-time feasibility remains a critical and ongoing research challenge.

The main contributions of this paper are:

Graph guided convex corridor construction: A graph search guided convex lateral corridor construction method is proposed, where a rough collision-free path is explicitly used to define a spatially varying feasible region for optimization.


MPC-based path refinement: Within the constructed convex corridor, the path refinement problem is formulated as an MPC problem that penalizes the third-order spatial derivative of the lateral offset under feasibility constraints, thereby promoting smooth path profiles with reduced computational complexity.

The rest of this study is structured as follows: Section \ref{sec2} presents the methodology, Section \ref{sec3_new} describes the simulation setup and results while Section \ref{sec5} gives the conclusion.

%% file: 02_New_Method.tex
\section{Methodology} \label{sec2}

\begin{algorithm}[H]
\caption{Integrated Graph Search - MPC Path Planning}
\label{al1}
\small
\begin{algorithmic}[1]
\Require Map $\mathcal{M}$, obstacle set $\mathcal{O}$, initial state $x_0$
\Ensure Optimized trajectory $\{x_k\}_{k=0}^{N_p}$

\State Discretize $\mathcal{M}$ into grid with adaptive resolution $(d_s, d_l)$
\State Mark occupied cells based on obstacle set $\mathcal{O}$
\State Find rough path $\mathcal{P} = \{(s_i, l_i)\}$ using Dijkstra search

\State Construct convex lateral safety corridor 
$\mathcal{C}$

\State Initialize MPC state $x \leftarrow x_0$

\For{$k = 0$ to $N_p-1$}
    \State Formulate and solve MPC problem 
    \State Apply first control input $u_k$
    \State Update state $x_{k+1}$
\EndFor

\State \Return optimized trajectory $\{x_k\}$
\end{algorithmic}
\end{algorithm}


\begin{figure}[h!]
    \centering
    \includegraphics[width=1\linewidth]{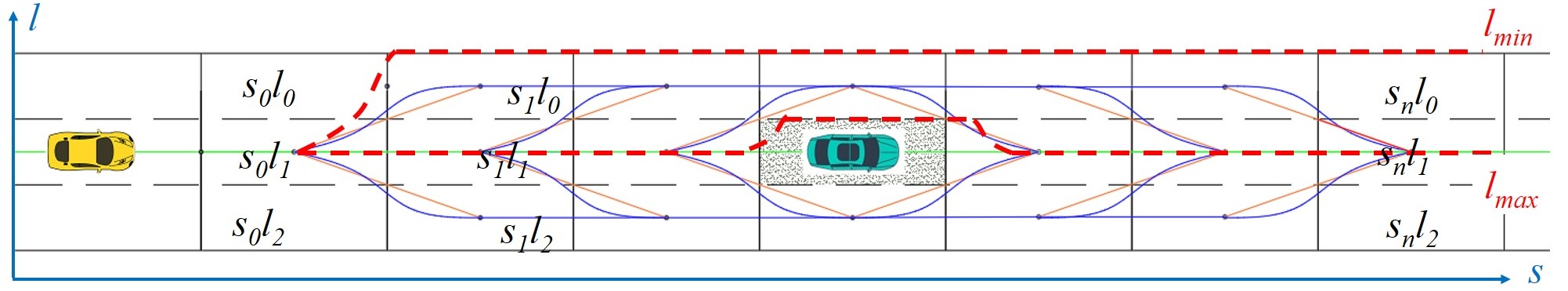}
    \caption{Path planning generation: from the global path (green), the map is discretized into cells and occupied cells are marked. A rough reference path (orange) is searched and subsequently smoothed by MPC (blue), satisfying the boundary lines (red dot line).}
    \label{fig2}
\end{figure}

Algorithm \ref{al1} illustrates the path planning flow in four steps. First, the map is discretized into cells, with target vehicles marked as occupied. Next, Dijkstra’s algorithm \cite{rachmawati2020analysis} is used to find the shortest path through the unoccupied cells. Based on this path, a safe corridor is built by expanding the surrounding convex space to form a feasible region. 
Finally, MPC computes the path by optimizing vehicle motion under smoothness and acceleration constraints.

\subsection{Grid map and Graph search}

In this study, we use the Frenet coordinate system \cite{meng2019decoupled}, \cite{zhang2024intelligent}, \cite{li2023quantitative} to represent vehicle motion relative to the road. The considered map is discretized into cells, the cells occupied by obstacles are marked as unsafe, as illustrated in Fig \ref{fig2}.


\subsubsection{Rough Path Generation}

A Dijkstra-based search is performed on a discretized grid to compute a safe path $(s,l)$ for the ego vehicle, where $s$ and $l$ denote the longitudinal and lateral coordinates in meters, respectively. The algorithm employs an adaptive resolution scheme for the grid construction, for complex maneuvers such as overtaking, the map is discretized with a high-resolution grid as:
\begin{equation}
    \begin{aligned}
    s(i) & = s_0 + i \cdot ds \\
    l(j) & = (n_{l,\text{left}} + 1 - j) \cdot dl
    \end{aligned}
\end{equation}
where $ds$ and $dl$ represent the fine longitudinal and lateral grid resolutions, $n_{l,\text{left}}$ is the number of lateral cells to the left of the reference line, while $i$ and $j$ denote the respective grid indices. 
Conversely, the resolution is adaptively reduced in simpler driving states, such as, in the cruise control mode $l(i) = l_0 + i \cdot dl$ with $dl = 0$ to save the computational cost.


\subsubsection{Obstacle Boundaries}
For each static obstacle, the longitudinal and lateral extents are computed as:
\begin{align}
s_{\min} &= \textit s_{obs} - \frac{\textit{obs}_L}{2} \qquad l_{\min} = \textit l_{obs} - \frac{\textit{obs}_W}{2} \\
s_{\max} &= \textit s_{obs} + \frac{\textit{obs}_L}{2}   \qquad l_{\max} = \textit l_{obs} + \frac{\textit{obs}_W}{2}
\end{align}
where $\textit s_{obs}$, $\textit l_{obs}$ are the obstacle center, $\textit{obs}_L$, $\textit{obs}_W$ are its length and width, which are detected by the sensors. 

\subsubsection{Occupancy of Grid Cells}
Each cell $(i,j)$ in the grid map is marked as occupied if it intersects with an obstacle:
\begin{equation} \label{eq14}
\begin{cases}
(s_i + \tfrac{\Delta s}{2} > s_{min}) \wedge \;(s_i - \tfrac{\Delta s}{2} < s_{max}) \\
(l_j + \tfrac{\Delta l}{2} > l_{min}) \wedge \; 
(l_j - \tfrac{\Delta l}{2} < l_{max}) \\
\end{cases}
\end{equation}

\subsubsection{Transition Cost Function}
Defining $(p_i,p_j)$ are the target grid cells that the vehicle intends to move to from the current cell $ i,j$. The cost for a transition $(i,j) \to (p_i,p_j)$ is:
\begin{equation}
C(i,j \to p_i,p_j) = C_b(j,p_j) + C_p(p_i,p_j)
\end{equation}

The path is designed such that the ego vehicle prioritises overtaking on the left. Therefore, the basic lane-change cost and the lateral penalty are defined as:
\begin{equation}
C_b(j,p_j) =
\begin{cases}
1, & p_j = j \quad \text{(go straight)} \\
1.1, & p_j < j \quad \text{(move left)} \\
1.5, & p_j > j \quad \text{(move right)}
\end{cases}
\end{equation}

\begin{equation}
C_p(i,j) =
\begin{cases}
10^5, & l(i,j) < l_0 - 0.1 \quad \text{(right overtake)} \\
0.5, & l(i,j) > l_0 + 0.1 \quad \text{(left overtake)} \\
0.1, & \text{otherwise (go straight)}
\end{cases}
\end{equation}

The cumulative distance is updated as:
\begin{equation}
d(p_i,p_j) = \min \big[d(p_i,p_j), \, d(i,j) + C(i,j \to p_i,p_j) \big]
\end{equation}
with $d(i,j)$ represents the accumulated minimum cost to reach the cell $(p_i,p_j)$.

The shortest path is obtained by backtracking from:
\begin{equation} \label{eq11}
j^* = \arg\min_j \text{dist}(n_s,j)
\end{equation}
with $n_s$ is the index that indicates the longitudinal position of the path endpoint. The set of all path segments that satisfy the equation (\ref{eq11}), the rough path is constructed as:
\begin{equation} \label{eq12}
(s,l) = \big(s(i_1,j_1),\dots,s(i_k,j_k); l(i_1,j_1),\dots,l(i_k,j_k) \big)
\end{equation}

\subsection{Determination of the Lateral Safety Corridor}
Given the path $(s,l)$ from equation (\ref{eq12}) and a set of obstacles positions, the lateral boundaries of the convex space are $l_{\min}(j), \quad l_{\max}(j), \quad j=1,\dots,n$ in Figure \ref{fig2}.

For obstacle $i$, the longitudinal influence region is:
\begin{align}
s_{\min}^{(i)} &= s_{obs}(i) - \frac{obs_L}{2}  \\
s_{\max}^{(i)} &= s_{obs}(i) + \frac{obs_L}{2}  
\end{align}

Mapping to indices corresponding to $s_{\min}^{(i)}$ and $s_{\max}^{(i)}$ as:
\begin{align}
idx_s^{(i)}  &= \arg\min_j | s_j - s_{\min}^{(i)} | \\ 
idx_e^{(i)} &= \arg\min_j | s_j - s_{\max}^{(i)} |
\end{align}

The avoidance side is determined by:
\begin{equation}
\bar{l}_{\text{path}}^{(i)} = \frac{l(idx_s^{(i)} ) + l(idx_e^{(i)})}{2}
\end{equation}
with $\bar{l}_{\text{path}}^{(i)}$ specifies the lateral center of the safety corridor that the path should follow to avoid collisions.
Update rules for the left and right avoidance:
\begin{align} \label{eq25b}
l_{\min}(j) = \max \big( l_{\min}(j), l_{obs}^{(i)} + \frac{obs_W}{2} \big)  \quad \text{if} \quad \bar{l}_{path}^{(i)} \geq l_{obs}^{(i)}\\ 
l_{\max}(j) = \min \big( l_{\max}(j), l_{obs}^{(i)} - \frac{obs_W}{2} \big)  \quad  \text{if} \quad \bar{l}_{path}^{(i)} < l_{obs}^{(i)}
\end{align}
for $j \in [idx_s^{(i)} , idx_e^{(i)}]$. Now, the task is finding the optimal path (s,l) with $l \in [l_{min}, l_{max}]$ such that the path satisfies constraints regarding safety and smoothness.

\subsection{MPC-Based Optimal Path Generation}
The previous studies \cite{li2023quantitative, 8242694, 6225063, 7963597} built the QP problem to find the optimal path. Unlike those, in this paper, the path planning problem is formulated as an MPC problem.

Let  $f(s_k)= l_k$, $f'(s_k) = \dot l_k$, $f''(s_k)=\ddot l_k$, $u_k=\dddot l_k$ represent the derivatives of $f$ with respect to the longitudinal coordinate $s$.
The path can be expanded at $s_k$ up to the third order using Taylor expansion as follows \cite{zhang2024intelligent}:
\begin{equation} \label{eq20}
\begin{aligned}
l_{k+1} &= l_k + \dot l_k\,\Delta s + \tfrac{1}{2} \ddot l_k\,\Delta s^2 + \tfrac{1}{6}u_k\,\Delta s^3\\[3pt]
\dot l_{k+1} &=  \dot l_k + \ddot l_k\,\Delta s + \tfrac{1}{2}u_k\,\Delta s^2\\[3pt]
\ddot l_{k+1} &= \ddot l_k + u_k\,\Delta s
\end{aligned}
\end{equation}

Here, $l_k$ is the lateral offset, $\dot l_k$ and $\ddot l_k$ are its first and second derivatives with respect to longitudinal distance $s$, and $u_k$ is the third-order spatial derivative of the lateral offset (control input). 
The lateral motion of the ego vehicle along the reference path is discretized with sampling step $\Delta s$. 

Defining $N_p=20$ as the prediction horizon, for each prediction step $k=0,\dots,N_p$, the state and input vectors are: 
\begin{equation}
x_k = [\,l_k \ \dot l_k \ \ddot l_k\,]^\top, \qquad
u_k = \dddot l_k
\end{equation}

The dynamics in equation \eqref{eq20} can be written as the discrete-time state-space model:
\begin{equation} \label{eq22}
x_{k+1} = A x_k + B u_k
\end{equation}
with
\begin{equation} \label{eq23}
A =
\begin{bmatrix}
1 & \Delta s & \tfrac12 \Delta s^2\\
0 & 1 & \Delta s\\
0 & 0 & 1
\end{bmatrix},
\qquad
B =
\begin{bmatrix}
\tfrac16 \Delta s^3\\
\tfrac12 \Delta s^2\\
\Delta s
\end{bmatrix}
\end{equation}

The MPC problem is formulated as:
\begin{equation} \label{eq24}
\begin{aligned}
\min_{\substack{x_1, \dots, x_{N_p}, \\ u_0, \dots, u_{N_p-1}}} \;
& \sum_{i=0}^{N_p-1} 
\Big[
(x_i - x_{\text{ref},i})^\top Q (x_i - x_{\text{ref},i})
+ u_i^\top R u_i
\Big] \\
& \quad + (x_{N_p} - x_{\text{ref},N_p})^\top P (x_{N_p} - x_{\text{ref},N_p})
\end{aligned}
\end{equation}

\text{subject to: } \quad
\begin{align}
& x_{i+1} = A x_i + B u_i, \quad \forall i = 0, \dots, N_p - 1 \label{eq25} \\
& u_{\min} \le u_i \le u_{\max}, \quad \forall i = 0, \dots, N_p - 1 \label{eq26} \\
& A_{\text{sub}} x_i \le b_{\text{sub}}(i), \quad \forall i = 1, \dots, N_p \label{eq27} \\
& \Delta_{\min} \le G(x_{i+1} - x_i) \le \Delta_{\max}, \quad \forall i = 0, \dots, N_p-1 \label{eq28} \\
& x_0 = x_{\text{start}} \label{eq29}
\end{align}

In this formulation, the matrices $A, B$ in equation (\ref{eq25}) are presented in equation (\ref{eq23}).
 The parameters $(u_{\min}= -0.05 ~\mathrm{m}^{-2}, u_{\max}= 0.05~\mathrm{m}^{-2})$ in equation (\ref{eq26}) denote the lower and upper bounds of the control input. The matrices $A_{\text{sub}}, b_{\text{sub}}(i)$ in equation (\ref{eq27}) represent the convex corridor constraint, which will be detailed hereinafter. 
The constraint in equation (\ref{eq28}) imposes bounds on the variation of the first- and second-order spatial derivatives of the lateral offset between consecutive prediction steps to ensure smooth path transitions:
\begin{equation}
G = 
\begin{bmatrix}
0 & 1 & 0 \\
0 & 0 & 1
\end{bmatrix},
\
\Delta_{\max} = 
\begin{bmatrix}
\Delta_{\dot{l}}^{\max} \\
\Delta_{\ddot{l}}^{\max}
\end{bmatrix},
\
\Delta_{\min} = -\Delta_{\max}
\end{equation}
with $(\Delta_{\dot{l}}^{\max}$ = 0.2, $\Delta_{\ddot{l}}^{\max}$ =3.5$~\mathrm{m}^{-1}$) denoting the maximum allowed variations of the first- and second-order spatial derivatives, respectively.
Finally, $x_{\text{start}} \in \mathbb{R}^{3 \times 1}$ in equation (\ref{eq29}) gives the initial condition of the ego car. 
The reference state $x_{\text{ref},i} \in \mathbb{R}^{3 \times 1}$ for time step $i$ in the cost function in equation (\ref{eq24}) is given by:
\begin{align}
x_{\text{ref},i} & =
\begin{bmatrix}
\frac{l_{\min} + l_{\max}}{2} \quad  0 \quad  0
\end{bmatrix}^{\!\top},
\quad \forall i = 0, \dots, N_p - 1 \\
x_{\text{ref},N_p} & = x^*
\end{align}
with $x^*= [l_{end}, \dot l_{end}, \ddot l_{end}]^T$ the desired terminal state. The weighting matrices are given by:
\begin{equation}
Q = \operatorname{diag}\{w_l, w_{\dot{l}}, w_{\ddot{l}}\}, 
\quad P = w_{\text{end}} I_3, \quad R = w_u
\end{equation}

where the weight factors $w_l$, $w_{\dot{l}}$, $w_{\ddot{l}}$, $w_{\text{end}}$, and $w_u$ are  selected empirically. Specifically,  higher weights are assigned to the first- and second-order spatial derivatives $w_{\dot{l}}=1500$ and $w_{\ddot{l}}=200$, while lower weights for the lateral position $w_l=3$, $w_{\text{end}}=15$, control input $w_u=20$ to refine the path and regularize the control effort.


\textbf{ Convex corridor constraints}: For each step $i=1,\dots,N_p$, the ego vehicle must remain inside the convex safe corridor:
\begin{equation}
A_{\text{sub}}\, x_i \le b_{\text{sub}}(i)
\end{equation}
with:
\begin{align}
A_{\text{sub}} =
\begin{bmatrix}
 1 &  d_1 & 0\\
 1 &  d_1 & 0\\
 1 & -d_2 & 0\\
 1 & -d_2 & 0\\[3pt]
-1 & -d_1 & 0\\
-1 & -d_1 & 0\\
-1 &  d_2 & 0\\
-1 &  d_2 & 0
\end{bmatrix},
\
b_{\text{sub}}(i) =
\begin{bmatrix}
l_{\max}(f_i) - \tfrac{w}{2}\\
l_{\max}(f_i) + \tfrac{w}{2}\\
l_{\max}(b_i) - \tfrac{w}{2}\\
l_{\max}(b_i) + \tfrac{w}{2}\\[3pt]
-\,l_{\min}(f_i) + \tfrac{w}{2}\\
-\,l_{\min}(f_i) - \tfrac{w}{2}\\
-\,l_{\min}(b_i) + \tfrac{w}{2}\\
-\,l_{\min}(b_i) - \tfrac{w}{2}
\end{bmatrix}
\end{align}

where $f_i $ and $b_i$ are the  range of prediction forward and backward influence indices:
\[
f_i = \min\!\big(i+\lceil d_1/\Delta s\rceil,N_p\big),\quad
b_i = \max\!\big(i-\lceil d_2/\Delta s\rceil,1\big).
\]

By using the solver "mpcActiveSetSolver" in MATLAB to solve this problem under the constraints, a convex, smooth and dynamically feasible path is obtained.

%% file: 03_Simulation_results.tex
\section{Simulation and results}\label{sec3_new}

In this study, the proposed algorithm is implemented using MATLAB/Simulink R2021b \cite{mathworks2021matlab} for control development, IPG CarMaker 11.1.1 \cite{IPG} and the BMW 5-Series vehicle model with calibrated parameters from \cite{bui2023lateral} for high-fidelity vehicle dynamics simulation. All simulations and experiments are conducted on a workstation running Microsoft Windows 11 Pro. The hardware platform is an HP ZBook Fury 15.6" G8 Mobile Workstation, equipped with an 11th Gen Intel® Core™ i7-11800H processor (2.30 GHz, 8 cores, 16 threads) and 32 GB of DDR4 RAM. 
\pagestyle{fancy}
\fancyhf{} 
\thispagestyle{fancy}

To evaluate the performance of the path planning algorithm, we constructed three simulation scenarios: (1) the AV overtakes a single target car (TG) on a straight road; (2) the AV overtakes a single TG on a curved road; and (3) the AV drives on a straight road with multiple target cars.
For comparisons, we employ the algorithm used in a previous study \cite{li2023quantitative}, which uses a discretization approach with sampled points combined with polynomial fitting and the QP method. The comparison aspects are lateral acceleration, jerk, curvature, steering angle and computational cost in scenarios 1 and 2.
The video simulation results can be found at \textbf{link} \cite{Bui2026Video}.

\subsection{Scenario 1:  Straight road – single target vehicle} 
\begin{figure} [h!]
    \centering
    \includegraphics[width=1\linewidth]{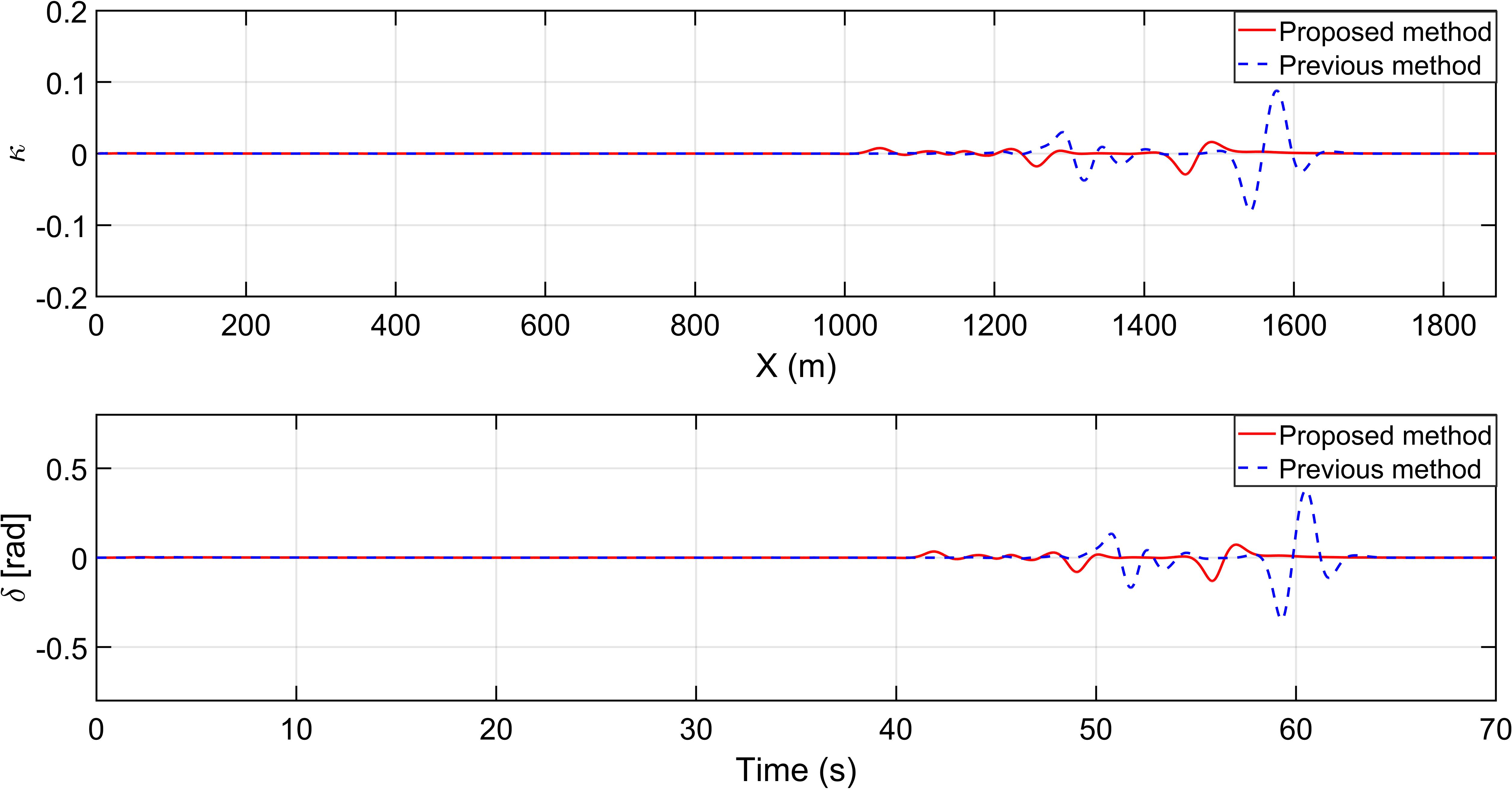}
    \caption{Curvature and steering angle on the straight road with one TG}
    \label{fig3}
\end{figure}

\begin{figure}[h!]
    \centering
    \includegraphics[width=1\linewidth]{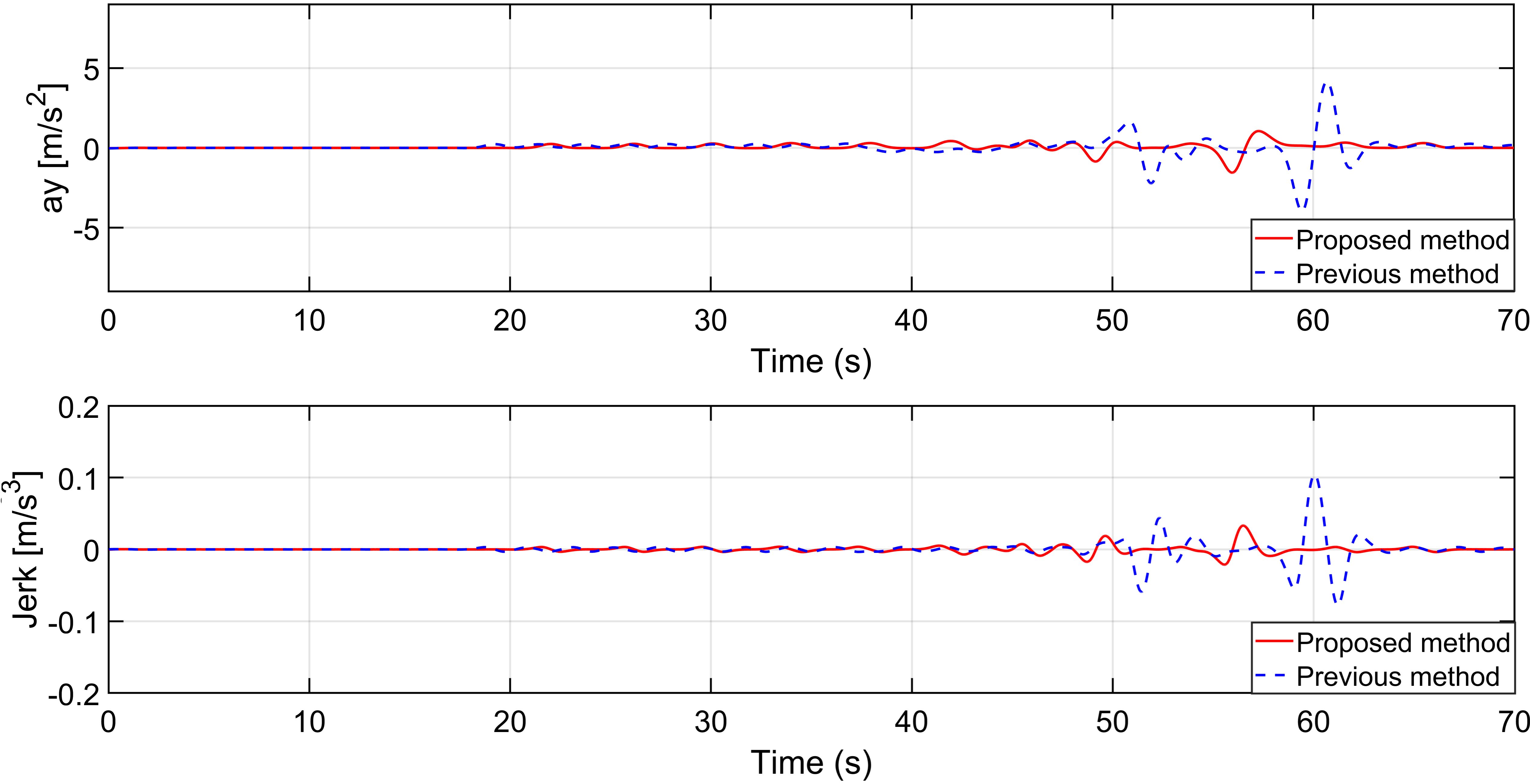}
    \caption{Lateral acceleration and jerk on the straight road with one TG}
    \label{fig4}
\end{figure}

The AV drives at a speed of {110}{km/h} and overtakes a TG moving at {90}{km/h} on a straight road. The simulation duration for this scenario is set at {75}{s}.
The lateral acceleration, steering angle, curvature and jerk of the ego car are shown in Figure \ref{fig3}, \ref{fig4} and Table \ref{tab2}.


\subsection{Scenario 2: Curved road – single target vehicle} 
\begin{figure}[h!]
    \centering
    \includegraphics[width=1\linewidth]{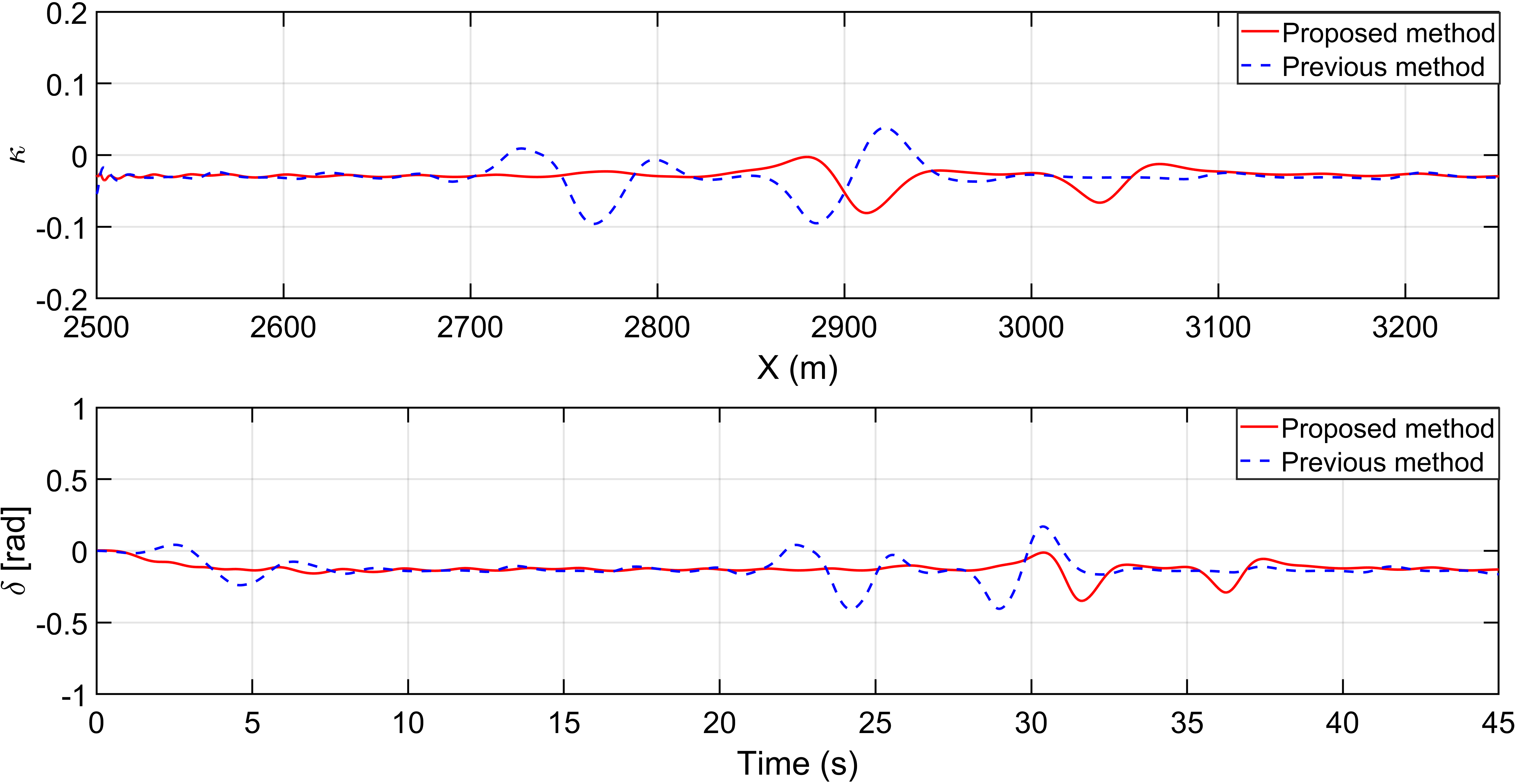}
    \caption{Curvature and steering angle on the curved road with one TG}
    \label{fig5}
\end{figure}

\begin{figure}[h!]
    \centering
    \includegraphics[width=1\linewidth]{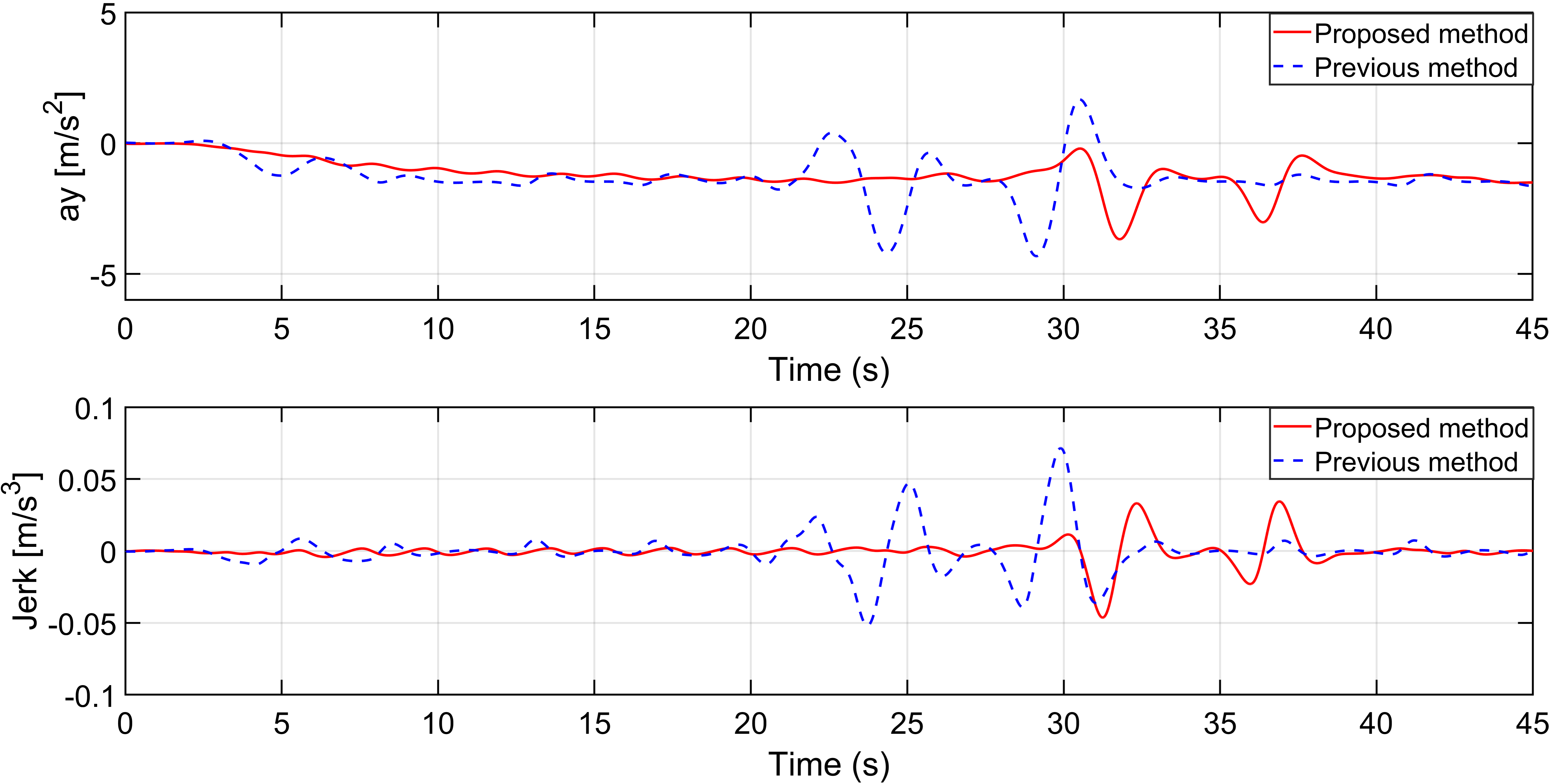}
    \caption{Lateral acceleration and jerk on the curved road with one TG}
    \label{fig6}
\end{figure}

The AV moves at {100}{km/h} on a curved road with a radius of R = {500}{m}, overtaking a target car driving at {80}{km/h}. 
The simulation setting time is 45{s}. 
The obtained results related to the lateral acceleration, steering angle, curvature, jerk and the computation cost when the ego car overtakes the TG on the curved road are presented in the Figure \ref{fig5}, \ref{fig6} and Table \ref{tab2}.\\
As illustrated in Table \ref{tab2}, for the scenarios 1 and 2, where the ego vehicle overtakes a single TG, the proposed algorithm exhibits smaller values of lateral acceleration, jerk, curvature and steering angle compared to the previous study \cite{li2023quantitative}. Specifically, the maximum values of these parameters for the proposed algorithm are $1.028\mathrm{m/s^2}$, $0.034\mathrm{m/s^3}$, $0.016\mathrm{m^{-1}}$ and $0.072\mathrm{rad}$, respectively, whereas the corresponding values in \cite{li2023quantitative} are $4.140\mathrm{m/s^2}$, $0.103\mathrm{m/s^3}$, $0.087\mathrm{m^{-1}}$ and $0.378\mathrm{rad}$. Similarly, the minimum values of lateral acceleration, jerk, curvature and steering angle in the proposed algorithm are $-3.491\mathrm{m/s^2}$, $-0.046\mathrm{m/s^3}$, $-0.080\mathrm{m^{-1}}$ and $-0.349\mathrm{rad}$, compared to $-4.322\mathrm{m/s^2}$, $-0.077\mathrm{m/s^3}$, $-0.096\mathrm{m^{-1}}$ and $-0.407\mathrm{rad}$ in \cite{li2023quantitative}. These results indicate the improved smoothness performance of the proposed algorithm.

The computational advantage of the proposed method is further underscored by the reduction in the average computation time per step, which decreased from $0.0021\text{s}$ in the previous study to $0.0017\text{s}$. This reduction time plays a pivotal role in maintaining vehicle stability and minimizing algorithmic latency between sensing and actuation during high-speed maneuvers. 
The observed enhancement in efficiency arises from explicitly utilizing the Dijkstra-generated rough path to construct a spatially varying convex safety corridor, whereby the complex problem of obstacle avoidance is transformed into a series of continuous feasibility constraints. 
This structural simplification allows the mpcActiveSetSolver to converge more rapidly compared to the polynomial fitting and traditional QP approaches used in \cite{li2023quantitative}. Furthermore, the adoption of a prediction horizon ($N_p = 20$) instead of solving for the entire path simultaneously as a QP problem, significantly reduces the dimensionality of the optimization vector, thereby lowering the cumulative computational overhead while ensuring sufficient look-ahead capability for smooth path refinement. 
Consequently, the proposed method requires only  $34.847\mathrm{s}$ and $18.213\mathrm{s}$ for the path planning module alone to execute scenarios 1 and 2, respectively. In contrast, the method reported in \cite{li2023quantitative} requires $48.452\mathrm{s}$ and $25.843\mathrm{s}$ for the path planning module for the same scenarios. 
Therefore, the proposed algorithm not only achieves a $28.08\%$ to $29.52\%$ reduction in total execution time in scenario 1 and scenario 2 but also supports real-time applicability under the tested simulation conditions.



\subsection{Scenario 3: Straight road – multiple target vehicles} 
\begin{figure} [h!]
    \centering
    \includegraphics[width=1\linewidth]{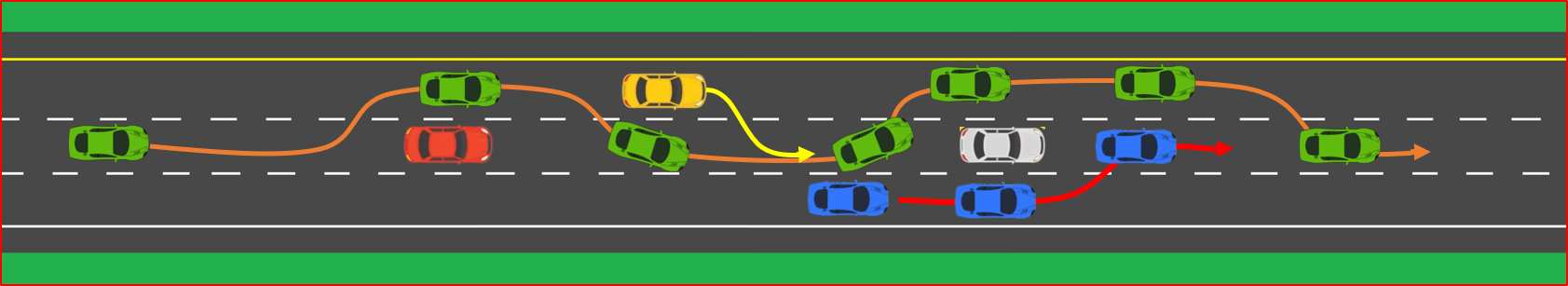}
    \caption{Scenario 3: Straight road – multiple target vehicles}
    \label{fig14}
\end{figure}

The AV drives at $70\mathrm{km/h}$ on a straight road with multiple target vehicles, see Figure \ref{fig14}.
Initially, two vehicles are present: TG1 (red) in the middle lane and TG2 (yellow) in the left lane, both moving at $50\mathrm{km/h}$. Since the safety distance between TG1 and TG2 is sufficient, the AV successfully overtakes TG1.
TG2 then performs a deceleration and lane-change manoeuvre into the middle lane.
Subsequently, the AV encounters TG3 (white) in the middle lane and TG4 (blue) in the right lane, both driving at $50\mathrm{km/h}$.
While overtaking TG3, TG4 suddenly accelerates, cuts into TG3’s lane, and continues in the same lane. 
The AV then re-plans path and overtakes both TG3 and TG4 safely.

\begin{table}[h!]
\centering
\caption{Comparison between Proposed Algorithm and Previous Study}
\label{tab2}
\resizebox{0.99\linewidth}{!}{%
\begin{tabular}{|c|c|c|c|c|c|c|}
\hline
{\textbf{Features}} & \textbf{Scenario} & \textbf{Algorithm}  & \textbf{Max} & \textbf{Min} & \textbf{Unit} \\ 
\hline
\cline{3-6}
 & {\textbf{1}} & Proposed Algorithm &  1.028 & -1.545&$m/s^2$ \\ 
\cline{3-6}
{\textbf{Lateral }}&     & Previous study    &   4.140& -3.966&$m/s^2$ \\ 
\cline{2-6}
 {\textbf{acceleration}}  & {\textbf{2}} & Proposed Algorithm & -0.210 & -3.491&$m/s^2$ \\ 
\cline{3-6}
            &     & Previous study    & 1.666  & -4.322&$m/s^2$ \\  \hline
& {\textbf{1}} & Proposed Algorithm &  0.033 & -0.021&$m/s^3$\\ 
\cline{3-6}
{\textbf{Jerk}} &     & Previous study    &   0.103 & -0.077 &$m/s^3$\\ 
\cline{2-6}
                & {\textbf{2}} & Proposed Algorithm & 0.034 &-0.046 &$m/s^3$\\ 
\cline{3-6}
 &     & Previous study    &  0.070  & -0.051&$m/s^3$\\  \hline
{\textbf{}} & {\textbf{1}} & Proposed Algorithm &  0.016 & -0.029&$m^{-1}$\\ 
\cline{3-6}
{\textbf{Curvature}}&     & Previous study    &  0.087 & -0.080 &$m^{-1}$\\ 
\cline{2-6}
                & {\textbf{2}} & Proposed Algorithm & -0.003 & -0.080&$m^{-1}$\\ 
\cline{3-6}
                &     & Previous study  & 0.038  &-0.096&$m^{-1}$\\  \hline
 & {\textbf{1}} & Proposed Algorithm &  0.072 & -0.130 &$rad$\\ 
\cline{3-6}
{\textbf{Steering}} &     & Previous study    & 0.378 & -0.346 &$rad$\\ 
\cline{2-6}
{\textbf{angle}}  & {\textbf{2}} & Proposed Algorithm &-0.014 &-0.349&$rad$\\ 
\cline{3-6}
                &     & Previous study    &0.17 &-0.407 &$rad$\\  \hline

\multicolumn{3}{|c|}{\textbf{Simulation time setup in scenario 1}} & \multicolumn{2}{c|}{\text{75}} & $s$ \\ 
\hline
\multicolumn{3}{|c|}{\textbf{Simulation time setup in scenario 2}} & \multicolumn{2}{c|}{\text{45}} & $s$ \\ 
\hline
\multicolumn{2}{|c|}{\textbf{Average computation  }} & {Proposed Algorithm} & \multicolumn{2}{|c|}{0.0017}  &$s$ \\ 
\cline{3-6}
\multicolumn{2}{|c|}{\textbf{time per step}} & {Previous study} & \multicolumn{2}{|c|}{0.0021}  &$s$ \\ 
\hline
\multicolumn{2}{|c|}{\textbf{Execution time of path}} & {Proposed Algorithm} & \multicolumn{2}{|c|}{34.847}  &$s$ \\ 
\cline{3-6}
\multicolumn{2}{|c|}{\textbf{planning in scenario 1}} & {Previous study} & \multicolumn{2}{|c|}{48.452} & $s$ \\ 
\hline
\multicolumn{2}{|c|}{\textbf{Execution time of path}} & {Proposed Algorithm} & \multicolumn{2}{|c|}{18.213} &$s$ \\ 
\cline{3-6}
\multicolumn{2}{|c|}{\textbf{planning in scenario 2}} & {Previous study} & \multicolumn{2}{|c|}{25.843} &$s$ \\ 
\hline
\end{tabular}
}
\end{table}

\begin{figure}[h!]
    \centering
    \includegraphics[width=1\linewidth]{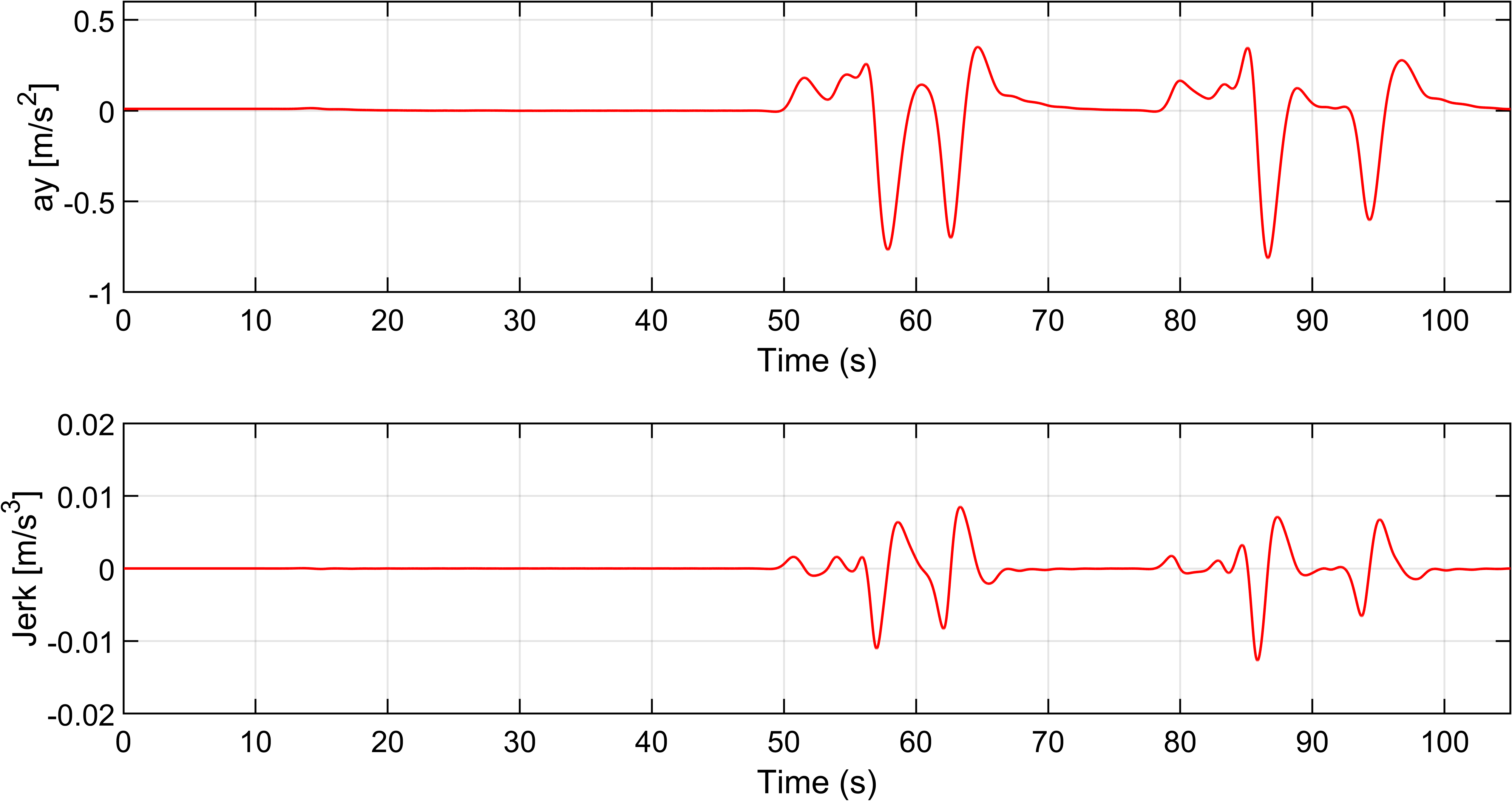}
    \caption{Lateral acceleration and jerk on the straight road with multiple target cars}
    \label{fig15}
\end{figure}

\begin{figure}[h!]
    \centering
    \includegraphics[width=1\linewidth]{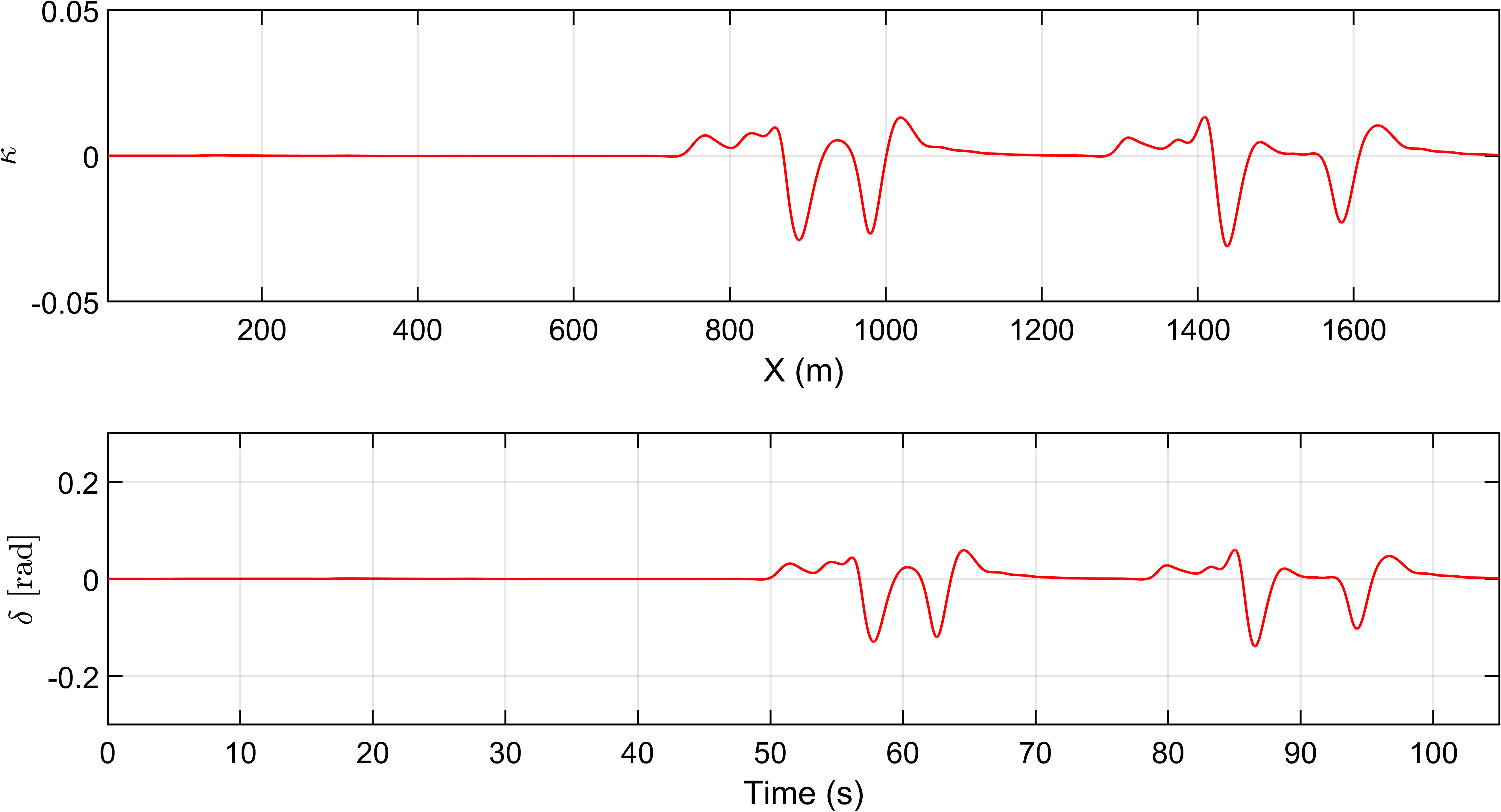}
    \caption{Curvature and steering angle on the straight road with multiple target cars}
    \label{fig16}
\end{figure}




Figure \ref{fig15} and \ref{fig16} show the lateral acceleration, steering angle, curvature and jerk when the ego car drives on the straight road with multiple target cars.  Specifically, the lateral acceleration is bounded between $-0.812$ and 0.341$\mathrm{m/s^2}$, and the resulting steering angle is kept between $-0.128$ and $0.06$$\mathrm{rad}$. 
Similarly, the curvature spans from $-0.028$ to $0.013$$\mathrm{m^{-1}}$ and the jerk is small, ranging from $-0.013$ to $0.008$$\mathrm{m/s^3}$. These values confirm that the proposed algorithm can consistently output a smooth and  feasible path that ensures driving comfort and stability.

%% file: 05_CONCLUSION.tex
\section{Conclusion} \label{sec5}
This paper introduced a new path-planning algorithm that integrates graph search, the Dijkstra algorithm and MPC. The proposed method discretizes the obstacle avoidance space according to varying driving environments, formulates constraints, and constructs a convex feasible region to represent vehicle–obstacle interactions. 
Within this region, an objective function is established and solved based on MPC algorithm to derive the optimal avoidance trajectory. 
This formulation enables efficient and reliable obstacle avoidance across a wide range of scenarios on both straight and curved roads. Simulation results demonstrate that the proposed method adapts effectively to diverse driving conditions, generating smooth and dynamically feasible paths while achieving a lower computational cost compared to existing approaches.

Future work will focus on further optimizing and testing the proposed algorithm under more complex traffic scenarios.


\thispagestyle{empty} 

%% file: Refers.bib
@electronic{Bui2026Video,
  author       = {Duc-Tien Bui},
  title        = {Integrated Graph Search and Model Predictive Control for Smooth and Efficient Path Planning in Autonomous Vehicles},
  year         = {2026},
  note         = {Available: \url{https://www.youtube.com/watch?v=0SeYLDn2E6E}, Accessed: Apr. 8, 2026},
  organization = {YouTube}
}

@article{zhang2023bi,
  title={Bi-AM-RRT*: A fast and efficient sampling-based motion planning algorithm in dynamic environments},
  author={Zhang, Ying and Wang, Heyong and Yin, Maoliang and Wang, Jiankun and Hua, Changchun},
  journal={IEEE Transactions on Intelligent Vehicles},
  volume={9},
  number={1},
  pages={1282--1293},
  year={2023},
  publisher={IEEE}
}

@manual{mathworks2021matlab,
  title        = {MATLAB and Simulink Release R2021b},
  author       = {{The MathWorks, Inc.}},
  organization = {The MathWorks, Inc.},
  address      = {Natick, Massachusetts, United States},
  year         = {2021},
  note         = {Available: https://www.mathworks.com/}
}

@inproceedings{yijing2018local,
  title={Local path planning of autonomous vehicles based on A* algorithm with equal-step sampling},
  author={Yijing, WANG and Zhengxuan, LIU and Zhiqiang, ZUO and Zheng, Ll},
  booktitle={2018 37th Chinese Control Conference (CCC)},
  pages={7828--7833},
  year={2018},
  organization={IEEE}
}

@article{hart1968formal,
  title={A formal basis for the heuristic determination of minimum cost paths},
  author={Hart, Peter E and Nilsson, Nils J and Raphael, Bertram},
  journal={IEEE transactions on Systems Science and Cybernetics},
  volume={4},
  number={2},
  pages={100--107},
  year={1968},
  publisher={IEEE}
}

@article{fan2018baidu,
  title={Baidu apollo em motion planner},
  author={Fan, Haoyang and Zhu, Fan and Liu, Changchun and Zhang, Liangliang and Zhuang, Li and Li, Dong and Zhu, Weicheng and Hu, Jiangtao and Li, Hongye and Kong, Qi},
  journal={arXiv preprint arXiv:1807.08048      }      ,   
  year={2018}
}

@ARTICLE{11404227,
  author={Li, Zheng and Wang, Yijing and Zuo, Zhiqiang and Zhao, Rui and Hu, Chuan and Shi, Yang},
  journal={IEEE Transactions on Intelligent Transportation Systems}, 
  title={Model Predictive Control-Based Trajectory Optimization for Autonomous Vehicles Using Risk-Aware Corridors}, 
  year={2026},
  volume={},
  number={},
  pages={1-16},
  doi={10.1109/TITS.2026.3664712}}

@article{katrakazas2015real,
  title={Real-time motion planning methods for autonomous on-road driving: State-of-the-art and future research directions},
  author={Katrakazas, Christos and Quddus, Mohammed and Chen, Wen-Hua and Deka, Lipika},
  journal={Transportation Research Part C: Emerging Technologies},
  volume={60},
  pages={416--442},
  year={2015},
  publisher={Elsevier}
}

@article{nezami2024safe,
  title={Safe Control Architecture via Model Predictive Control},
  author={Nezami, Maryam and Nguyen, Ngoc Thinh and M{\"a}nnel, Georg and Kensbock, Robin and Abbas, Hossam Seddik and Schildbach, Georg},
  journal={IEEE Transactions on Control Systems Technology},
  year={2024},
  publisher={IEEE}
}

@article{nguyen2023risk,
  title={Risk-informed decision-making and control strategies for autonomous vehicles in emergency situations},
  author={Nguyen, Hung Duy and Choi, Mooryong and Han, Kyoungseok},
  journal={Accident Analysis \& Prevention},
  volume={193},
  pages={107305},
  year={2023},
  publisher={Elsevier}
}

@inproceedings{rachmawati2020analysis,
  title={Analysis of Dijkstra’s algorithm and A* algorithm in shortest path problem},
  author={Rachmawati, Dian and Gustin, Lysander},
  booktitle={Journal of Physics: Conference Series},
  volume={1566},
  number={1},
  pages={012061},
  year={2020},
  organization={IOP Publishing}
}

@INPROCEEDINGS{9590740,
  author={Ren, Jing and Huang, Xishi},
  booktitle={2021 IEEE 4th International Conference on Information Systems and Computer Aided Education (ICISCAE)}, 
  title={Dynamic Programming Inspired Global Optimal Path Planning for Mobile Robots}, 
  year={2021},
  volume={},
  number={},
  pages={461-465},
  doi={10.1109/ICISCAE52414.2021.9590740 }}

@article{abdallaoui2022thorough,
  title={Thorough review analysis of safe control of autonomous vehicles: path planning and navigation techniques},
  author={Abdallaoui, Sara and Aglzim, El-Hassane and Chaibet, Ahmed and Krib{\`e}che, Ali},
  journal={Energies},
  volume={15},
  number={4},
  pages={1358},
  year={2022},
  publisher={MDPI}
}

@INPROCEEDINGS{7963597,
  author={Liu, Changliu and Lin, Chung-Yen and Wang, Yizhou and Tomizuka, Masayoshi},
  booktitle={2017 American Control Conference (ACC)}, 
  title={Convex feasible set algorithm for constrained trajectory smoothing}, 
  year={2017},
  volume={},
  number={},
  pages={4177-4182},
  doi={10.23919/ACC.2017.7963597 }}

@INPROCEEDINGS{6225063,
  author={Wenda Xu and Junqing Wei and Dolan, John M. and Huijing Zhao and Hongbin Zha},
  booktitle={2012 IEEE International Conference on Robotics and Automation}, 
  title={A real-time motion planner with trajectory optimization for autonomous vehicles}, 
  year={2012},
  volume={},
  number={},
  pages={2061-2067},
  doi={10.1109/ICRA.2012.6225063}}

@ARTICLE{8242694,
  author={Lim, Wonteak and Lee, Seongjin and Sunwoo, Myoungho and Jo, Kichun},
  journal={IEEE Transactions on Intelligent Transportation Systems}, 
  title={Hierarchical Trajectory Planning of an Autonomous Car Based on the Integration of a Sampling and an Optimization Method}, 
  year={2018},
  volume={19},
  number={2},
  pages={613-626},
  doi={10.1109/TITS.2017.2756099}}

@ARTICLE{9211493,
  author={Wu, Chunjiang and Zhou, Shijie and Xiao, Licai},
  journal={IEEE Access}, 
  title={Dynamic Path Planning Based on Improved Ant Colony Algorithm in Traffic Congestion}, 
  year={2020},
  volume={8},
  number={},
  pages={180773-180783},
  doi={10.1109/ACCESS.2020.3028467}}

@inproceedings{chen2017constrained,
  title={Constrained iterative lqr for on-road autonomous driving motion planning},
  author={Chen, Jianyu and Zhan, Wei and Tomizuka, Masayoshi},
  booktitle={2017 IEEE 20th International conference on intelligent transportation systems (ITSC)},
  pages={1--7},
  year={2017},
  organization={IEEE}
}

@inproceedings{liu2017convex,
  title={Convex feasible set algorithm for constrained trajectory smoothing},
  author={Liu, Changliu and Lin, Chung-Yen and Wang, Yizhou and Tomizuka, Masayoshi},
  booktitle={2017 American Control Conference (ACC)},
  pages={4177--4182},
  year={2017},
  organization={IEEE}
}

@article{lim2018hierarchical,
  title={Hierarchical trajectory planning of an autonomous car based on the integration of a sampling and an optimization method},
  author={Lim, Wonteak and Lee, Seongjin and Sunwoo, Myoungho and Jo, Kichun},
  journal={IEEE Transactions on Intelligent Transportation Systems},
  volume={19},
  number={2},
  pages={613--626},
  year={2018},
  publisher={IEEE}
}

@article{hongyu2018improved,
  title={An improved artificial potential field model considering vehicle velocity for autonomous driving},
  author={Hongyu, HU and Chi, Zhang and Yuhuan, Sheng and Bin, Zhou and Fei, Gao},
  journal={IFAC-PapersOnLine},
  volume={51},
  number={31},
  pages={863--867},
  year={2018},
  publisher={Elsevier}
}

@article{gonzalez2015review,
  title={A review of motion planning techniques for automated vehicles},
  author={Gonz{\'a}lez, David and P{\'e}rez, Joshu{\'e} and Milan{\'e}s, Vicente and Nashashibi, Fawzi},
  journal={IEEE Transactions on intelligent transportation systems},
  volume={17},
  number={4},
  pages={1135--1145},
  year={2015},
  publisher={IEEE}
}

@article{bui2023lateral,
  title={Lateral control calibration and testing in a co-simulation framework for automated vehicles},
  author={Bui, Duc-Tien and Li, Hexuan and De Cristofaro, Francesco and Eichberger, Arno},
  journal={Applied Sciences},
  volume={13},
  number={23},
  pages={12898},
  year={2023},
  publisher={MDPI}
}

@article{reda2024path,
  title={Path planning algorithms in the autonomous driving system: A comprehensive review},
  author={Reda, Mohamed and Onsy, Ahmed and Haikal, Amira Y and Ghanbari, Ali},
  journal={Robotics and Autonomous Systems},
  volume={174},
  pages={104630},
  year={2024},
  publisher={Elsevier}
}

@article{zhou2022review,
  title={A review of motion planning algorithms for intelligent robots},
  author={Zhou, Chengmin and Huang, Bingding and Fr{\"a}nti, Pasi},
  journal={Journal of Intelligent Manufacturing},
  volume={33},
  number={2},
  pages={387--424},
  year={2022},
  publisher={Springer}
}

@article{zhang2024intelligent,
  title={Intelligent vehicle path based on discretized sampling points and improved cost function: A quadratic programming approach},
  author={Zhang, Chengtao and Xu, Weihang},
  journal={IEEE Access},
  volume={12},
  pages={24500--24515},
  year={2024},
  publisher={IEEE}
}

@article{liu2023path,
  title={Path planning techniques for mobile robots: Review and prospect},
  author={Liu, Lixing and Wang, Xu and Yang, Xin and Liu, Hongjie and Li, Jianping and Wang, Pengfei},
  journal={Expert Systems with Applications},
  volume={227},
  pages={120254},
  year={2023},
  publisher={Elsevier}
}

@article{meng2019decoupled,
  title={A decoupled trajectory planning framework based on the integration of lattice searching and convex optimization},
  author={Meng, Yu and Wu, Yangming and Gu, Qing and Liu, Li},
  journal={IEEE Access},
  volume={7},
  pages={130530--130551},
  year={2019},
  publisher={IEEE}
}

@article{li2023quantitative,
  title={Quantitative analysis of the impact of Baidu Apollo parameterization on trajectory planning in a critical scenario},
  author={Li, Hexuan and De Cristofaro, Francesco and Orucevic, Faris and Gu, Zhengguo and Eichberger, Arno},
  journal={Transportation Research Procedia},
  volume={73},
  pages={102--109},
  year={2023},
  publisher={Elsevier}
}

@TechReport{IPG,
  title       = {https://ipg-automotive.com/},
}
